\documentclass[conference]{IEEEtran}
\IEEEoverridecommandlockouts
\usepackage{cite}
\usepackage{amsmath,amssymb,amsfonts}
\usepackage{algorithmic}
\usepackage{graphicx}
\usepackage{textcomp}
\usepackage{xcolor}
\usepackage{todonotes}
\usepackage{array}
\usepackage[table]{xcolor}
\usepackage{booktabs,comment,hyperref,multirow}
\usepackage{subcaption}
\def\BibTeX{{\rm B\kern-.05em{\sc i\kern-.025em b}\kern-.08em
    T\kern-.1667em\lower.7ex\hbox{E}\kern-.125emX}}
\begin{document}

\title{CARB: A Dataset for Evaluating Context, Accent, Response, and Phonetic Ambiguity in ASR\thanks{\copyright~2026 IEEE. Personal use of this material is permitted. Permission from IEEE must be obtained for all other uses, in any current or future media, including reprinting/republishing this material for advertising or promotional purposes, creating new collective works, for resale or redistribution to servers or lists, or reuse of any copyrighted component of this work in other works. Accepted at IEEE SLT 2026.}}

\author{
\IEEEauthorblockN{Hamdan Al-Ali, Rufael Marew, Injy Hamed, Hanan Aldarmaki}
\IEEEauthorblockA{\textit{Mohamed bin Zayed University of Artificial Intelligence}\\
\textit{Abu Dhabi, UAE}}
}

\maketitle

\begin{abstract}
Automatic speech recognition (ASR) systems are increasingly deployed in interactive applications where they may be expected to operate in conversational settings with diverse accents and different contextual conditions. To enable controlled evaluations of how these conditions impact performance, %
 we introduce CARB, %
an English speech dataset designed to evaluate ASR systems under realistic deployment conditions, involving variations along five key dimensions: \underline{\textit{C}}ontext, \underline{\textit{A}}ccent, \underline{\textit{R}}esponse length and correctness, and \underline{\textit{P}}honetic ambiguity. 
The dataset contains 43 hours of speech, encompassing 2,244 context-answer pairs recorded in parallel by 138 speakers across 54 countries, and enriched with human  judgments. We benchmark eight SOTA ASR models on this dataset to demonstrate the concrete challenges and failure modes that CARB is designed to address. 
\end{abstract}


\section{Introduction}
Automatic speech recognition (ASR) models have achieved remarkable performance on benchmark datasets; yet, real-world deployment often involves conditions that are substantially more challenging than those represented in these benchmarks. As an example, the performance of a system trained on Librispeech \cite{panayotov2015librispeech}, which consists of standard native English read speech, could degrade sharply when applied to conversational or non-native accented speech. Furthermore, most state-of-the-art speech models are trained and validated on relatively long utterances, where the model could infer the target words from the linguistic context rather than acoustic cues, and the performance often degrades on short utterances with limited context. Evaluating ASR systems solely through aggregate recognition metrics on standard benchmarks provides limited insight into the behavior of the system in real-world applications. 

For example, in educational and reading assessment settings, %
accented but acceptable pronunciations may be misrecognized and erroneously flagged as errors, potentially causing frustrations and slowing down the learning process. Single-word utterances, whether accented or not, could also be challenging to recognize without additional context. Providing context to the ASR system could help disambiguate similar-sounding words, but may also lead to over-corrections and fail to detect pronunciation or reading errors. The line between acceptable speech and incorrect pronunciation is difficult to define and may be partially subjective. 

\begin{figure}
    \centering
    \includegraphics[width=\linewidth]{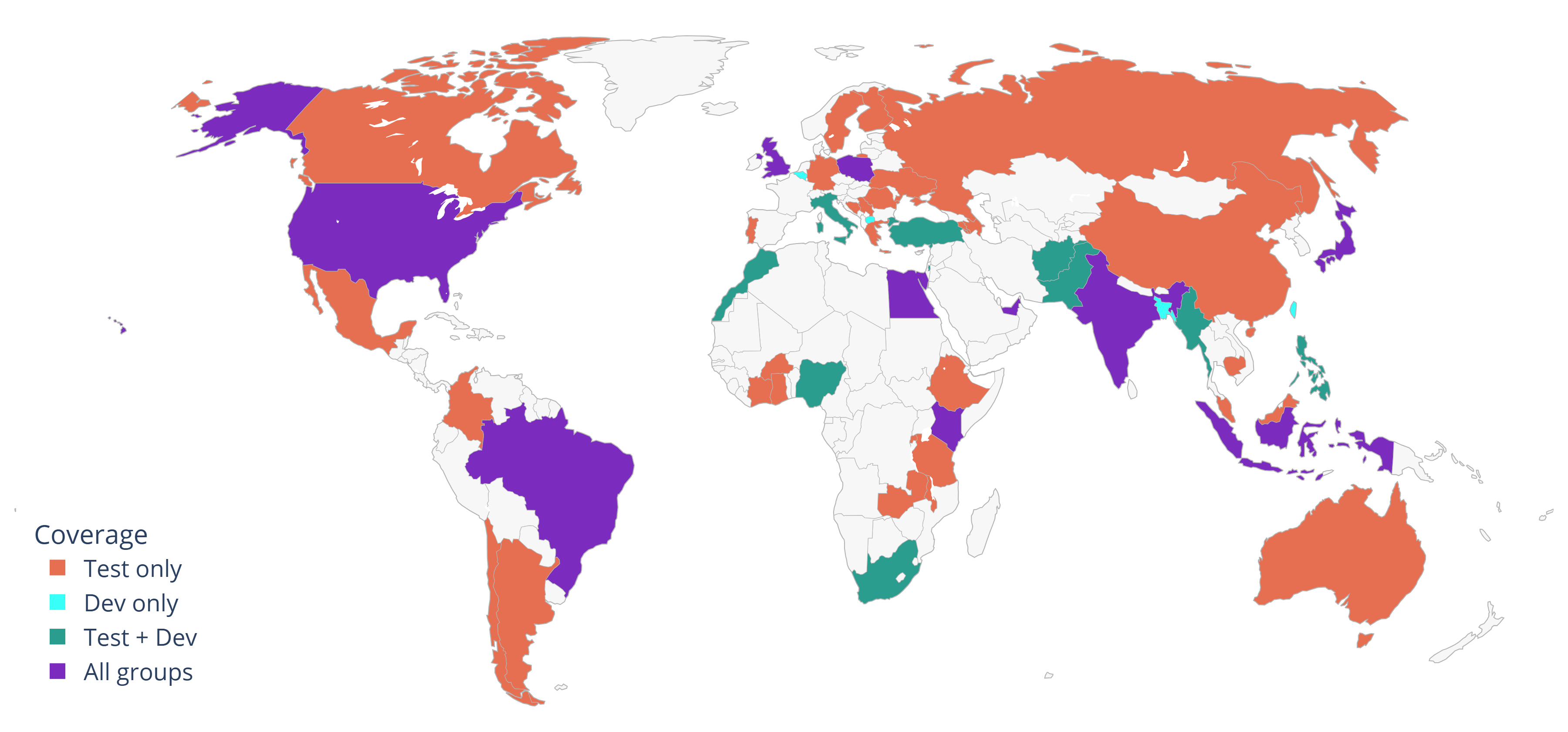}
    \caption{Geographic distribution of the countries represented in different splits of CARB.
    Dev and Test indicate countries represented only in the corresponding splits, while `All groups' indicates countries represented in all splits.
    }
    \label{fig:dataset_map}
\end{figure}

Existing English ASR datasets and benchmarks typically focus on individual aspects of the problem, such as accent variation (e.g., the Speech Accent Archive \cite{weinberger2011speech}), contextual speech recognition (e.g., Slidespeech \cite{10448079}),
pronunciation assessment (e.g., L2-ARCTIC ~\cite{zhao18b_interspeech}), or general transcription quality (e.g., Librispeech \cite{panayotov2015librispeech}, CommonVoice \cite{ardila2020common}). %
To enable more targeted and controlled evaluation, we introduce CARB\footnote{Example of a minimal pair that may cause phonetic ambiguity without context: Carp, a species of freshwater fish. Carb: short for carbohydrate.} (Context, Accent, Response, and Phonetic Ambiguity), a dataset designed to evaluate English ASR performance under realistic deployment conditions. CARB contains context-answer pairs spanning single-word, phrase-level, and sentence-level responses. The context-answer pairs are designed such that phonetically similar responses could be paired with the same context to test how models could correctly incorporate or overfit to context. The responses are recorded in parallel by speakers of diverse linguistic backgrounds (L1 and L2 speakers) to test how accent variations could impact performance under different conditions. In addition, we provide human assessment labels for a subset of responses to compare model performance against human acceptability judgments.  %
These design choices enable controlled evaluation of these dimensions of variation that reflect real-world ASR challenges and how they interact with each other and affect model recognition performance.

\begin{table}[ht]
\centering
\caption{Examples of scripted text from the CARB dataset. \textcolor{purple}{\textit{Incorrect, but phonetically similar responses.}}}
\label{tab:carb_examples}
\setlength{\tabcolsep}{2pt}
\rowcolors{2}{gray!10}{white}
\scalebox{0.9}{
\begin{tabular}{l>{\raggedright\arraybackslash}m{3cm}>{\raggedright\arraybackslash}m{3cm}cc}
\toprule
\textbf{U\_ID}&\textbf{Context}                             & \textbf{Answer}                                       & \textbf{Type} & \textbf{Correct?} \\\hline
\colorbox[HTML]{00CC66}{U0015}&What vehicle do you drive on the road?& Car& word& yes\\
U0016&What is the third person plural present of be?& Are& word & yes\\
U1923&What is a group classification called?& Species& word& yes\\
\colorbox[HTML]{CC33FF}{U1924}&What is a group classification called?& \textcolor{purple}{Spices}& word& no\\
U1597&What is extinction?& When a species no longer exists.& sentence& yes\\
\colorbox[HTML]{FF8C33}{U2024}&Where does an ocean fish live?& In the ocean & phrase& yes\\
\colorbox[HTML]{FFEE33}{U2025}&Where does an ocean fish live?& In the \textcolor{purple}{motion}& phrase& no\\
U0258&Name a type of energy.& Kinetic energy is energy of motion.& sentence& yes\\
\end{tabular}
}
\end{table}

In this paper, we describe CARB design and construction process and provide an overview of its content.  To demonstrate how the dataset can be used to conduct targeted evaluations of ASR systems, we benchmark eight  
state-of-the-art models that span diverse model architectures: auto-regressive encoder-decoder models, CTC-based encoder models, and RNN-T models. We demonstrate the performance on various dimensions, such as context availability, accent, and response length. %

The results reveal various limitations of state-of-the-art models, such as increased WER on shorter utterances and heavy accents, the uneven performance across demographies, and the complex effect of context on model predictions. CARB is publicly released under the CC-BY-NC 4.0 license,\footnote{\url{https://huggingface.co/datasets/potion-ae/CARB}} and the evaluation scripts are public,\footnote{\url{https://github.com/potion-ae/CARB}} to enable continual research.

\section{Dataset Design}

The central design principle of CARB is controlled variability. The dataset is constructed so that different evaluation settings can be created by selectively using or ignoring each of its five dimensions: 
context, accent, response length, response correctness, and phonetic ambiguity. This enables targeted assessment of ASR systems under different conditions. 
The dataset consists of context-response pairs. Contexts define a situation or prompt (e.g., a question or conversational turn), while responses are the primary speech signal intended for ASR evaluation. We present examples from CARB in Table~\ref{tab:carb_examples}, demonstrating variations present in the dataset.

\textbf{(1) Context:} Each item in CARB contains a textual context that represents the prompt or situation that is paired with the answer. 
Contexts are designed to narrow the semantic space of the response, without uniquely determining it, while allowing responses to differ in wording, length, and structure. 
The context enables analysis of contextual bias. %
For example, a prompt may strongly suggest a particular word, but the response may contain a phonetically similar word unrelated to the context. 
Depending on the application, leveraging context to resolve such ambiguity may be beneficial, or it may incorrectly override what the speaker actually said. CARB is designed to make such cases analyzable.

\textbf{(2) Accent:} Accent variation is a core component of CARB. The dataset is designed to include speakers from diverse geographic regions, covering 
North Africa, Sub-Saharan Africa, North America, Latin America and Caribbean, East Asia, South Asia, Southeast Asia, Western Asia, Eastern Europe and Balkans, Western Europe, Middle East, and Oceania. 
Some speakers are native (L1) speakers with different English dialects, and most are non-native (L2) speakers.  The dataset split is designed such that some accent groups are represented in training, while others are held-out for testing. This enables evaluation of ASR robustness across seen and unseen accents.

\begin{table}[]
\centering
\caption{Examples from CARB dataset metadata on the levels of speakers and recordings.}
\label{tab:carb_speaker_rec_example}
\rowcolors{2}{gray!10}{white}
\begin{minipage}{0.48\textwidth}
\centering
\scalebox{0.8}{
\begin{tabular}{llllp{0.8cm}ccc}
\toprule
& \multicolumn{4}{c}{\textbf{Self-Declared Attributes}} &\multicolumn{3}{c}{\textbf{Perceived degree of accent}} \\
\cmidrule(lr){2-7}
\cmidrule(lr){6-8}
 \multicolumn{1}{c}{\textbf{S\_ID}} & \multicolumn{1}{c}{\textbf{Gender}} & \multicolumn{1}{c}{\textbf{Language}} & \multicolumn{1}{c}{\textbf{Country}} & \textbf{Age} & \textbf{A1} & \textbf{A2} & \textbf{A3} \\
\hline
\colorbox[HTML]{38FFF8}{S068}& male & Japanese & Japan & 46 & Medium & Heavy & Medium \\
\colorbox[HTML]{ccff33}{S099}& female & Arabic & Egypt & 30 & Medium & Heavy & Heavy \\
\hiderowcolors
&&&&&&& \\
\end{tabular}
}
\subcaption{Speaker-level metadata.}
\label{tab:carb_speaker_example}
\end{minipage}
\hspace{0.01\textwidth}
\rowcolors{2}{gray!10}{white}
\begin{minipage}{0.45\textwidth}
\centering
\scalebox{0.78}{
\begin{tabular}{lllccccccc}
\toprule
& & & \multicolumn{3}{c}{\textbf{Needs Correction?}} & \multicolumn{3}{c}{\textbf{Error Words}} \\
\cmidrule(lr){4-6}\cmidrule(lr){7-9}
\multicolumn{1}{c}{\textbf{S\_ID}} & \multicolumn{1}{c}{\textbf{U\_ID}} & \multicolumn{1}{c}{\textbf{Answer}} & \textbf{A1} & \textbf{A2} & \textbf{A3} & \textbf{A1} & \textbf{A2} & \textbf{A3} \\
\hline
\colorbox[HTML]{38FFF8}{S068}& \colorbox[HTML]{FFEE33}{U2025}&In the motion & no & no & no & - & - & - \\
\colorbox[HTML]{38FFF8}{S068}& \colorbox[HTML]{FF8C33}{U2024} &In the ocean& yes & no & no & ocean & - & - \\
\colorbox[HTML]{ccff33}{S099}& \colorbox[HTML]{CC33FF}{U1924}&Spices & no & yes & no & - & spices & - \\
\colorbox[HTML]{ccff33}{S099}& \colorbox[HTML]{00CC66}{U0015}&Car & yes & yes & yes & car & car & car \\
\hiderowcolors
&&&&&&& \\
\end{tabular}
}
\subcaption{Recording-level metadata.}
\label{tab:carb_recording_example}
\end{minipage}
\end{table}

\textbf{(3) Response Length:} Responses in CARB are labeled by length category: single-word, phrase-level, or sentence-level. This design reflects real-world settings, where users may respond with varying utterance lengths to the same prompt. 
Single-word responses provide limited internal context for the recognizer. When the acoustic signal is ambiguous, such as between phonetically similar words like \textit{``cake''} and \textit{``kick''}, the model is more likely to rely on prior probabilities from learning history. Phrase-level and sentence-level responses provide additional internal context that can help disambiguate each target word. %

\textbf{(4) Response Correctness:} 
CARB includes both contextually correct and intentionally incorrect responses, as highlighted in Table \ref{tab:carb_examples}. Dataset entries are labeled, 
indicating whether the provided response appropriately answers the con or is unrelated to it. %
Including intentionally incorrect responses allows us to measure the impact of context on ASR performance, in particular to assess the degree of over-correction. 
This distinction is important for applications such as reading practice and  pronunciation assessment, %
where a system must 
maximize recognition performance while also detecting true pronunciation errors. 

\textbf{(5) Phonetic Ambiguity:} The responses in CARB include controlled phonetic ambiguity through sets of words that are intentionally similar in sound (minimal pairs). This makes it particularly challenging when combined with misleading context, which represents edge cases that actually arise in real-world educational applications. Table~\ref{tab:carb_examples} illustrates examples such as \textit{car}/\textit{are}, \textit{species}/\textit{spices}, and \textit{ocean}/\textit{motion}, where the disambiguating context may be contained within the response itself or provided through the paired context. This design allows CARB to support %
targeted analysis of phonetic confusion. \\%

\textbf{Together}, these dimensions allow CARB to be used flexibly across multiple evaluation settings. Researchers may evaluate the full dataset to study overall ASR robustness, or isolate specific subsets to examine the effects of context, accent variation, response length, response correctness, phonetic ambiguity, or any combination thereof. This controlled variability %
supports broad benchmarking and targeted analysis of specific 
ASR failure modes.

\section{Data Collection}
\noindent\textbf{Script Creation. }
The script for CARB was partially derived from educational applications. 
In these applications, users are prompted through images, text, or speech, and they provide spoken answers. We used some collected answers from proprietary applications as the basis for most of the script items, especially error cases observed in ASR outputs. For example, one item in CARB includes the answer \textit{The gail is leaping}, a case where the intended response was \textit{The girl is sleeping}.\footnote{Both the correct and incorrect variants are included in CARB.} In some cases, image-based contexts were converted into textual descriptions. 
For missing parts of the script, including answers without context and additional context-answer pairs, we used ChatGPT with GPT-5.5 Thinking\footnote{\url{https://openai.com/index/introducing-gpt-5-5/}} to generate candidate items. These were manually reviewed and edited. \\

\noindent\textbf{Participants}
 were recruited as adult L1 and L2 English speakers from diverse countries of origin. Participants were not required to be professional voice actors, and recordings were not collected in controlled studio conditions. This choice was intentional, as CARB is designed to reflect the average and likely speech variation and recording conditions. However, for every participant, a number of their recordings were analyzed to confirm the audio was intelligible, without extreme noises or mistakes. 
 Each participant self-reported their native language and country of origin. Speaker-level metadata is used to organize the dataset by accent groups. \\%

\noindent\textbf{Recording Protocol. }
The dataset consists of context- answer pairs, where the answer is the primary speech signal collected for ASR evaluation.
The script was divided into three dataset splits: train, dev, and test. For each split, we prepared separate collection sets for contexts and answers, resulting in six collection subsets: train-context, train-answer, dev-context, dev-answer, test-context, and test-answer. The primary objective is to benchmark model generalization to unseen speakers and accents, which is what is likely to be encountered in real-world applications. It is also possible to optimize models for specific target populations, or even personalize it to specific end-users. Keeping this in mind, we design the splits such that the test set includes parallel speech from both seen and unseen speakers and accent groups. The context script is not meant to be processed as the target of evaluation in this benchmark. Rather, it's meant to provide additional context for answers, such as a preceding conversational turn or a standardized system generated question or prompt. Therefore, we limit context recordings to few speakers.
Participants were instructed to record themselves reading each assigned utterance as a separate audio file and to read the answer exactly as written (including any intentionally added mistakes). For collecting read speech, we utilized the open-source toolkit Sentence-Recorder\footnote{\url{https://github.com/brianyan918/sentence-recorder}} described in \cite{yan2025cs}, which is a web-based tool for collecting read speech. We modified the toolkit to match our collection protocol. \\

\noindent\textbf{Consent and Compensation. }
Participants were hired through Upwork\footnote{\url{https://www.upwork.com/}}, and manually screened for geographic coverage.
Before recording, participants reviewed and signed an informed consent form describing the purpose of the study, the type of data collected, and the intended public release of the recordings. The consent form states that voice recordings will be released as part of an open research dataset and associated with anonymous speaker identifiers. Participants were compensated for approved recordings according to the agreed payment terms. \\

\noindent\textbf{Quality Control. }
To ensure %
recording quality and adherence to guidelines, participants first submitted a small set of trial recordings before completing the full assignment. Trial recordings were reviewed for audio usability, including background noise, microphone quality, in addition to script adherence and completeness. The same checks were applied to the full submission. 
Pronunciation correctness was not evaluated during this process and recordings were not rejected because of a speaker's accent even when it altered meaning. 
However, %
re-recording was requested when the submitted audio was unusable, %
or when the speaker clearly skipped, replaced, or omitted words from the provided text. \\%

\noindent\textbf{Accentedness Human Annotations. }
We %
annotated all speakers in the dataset for the degree of accented speech. Three native English language tutors from Canada, the USA, and the UK were hired to independently judge each speaker. %
The judges were instructed to listen to multiple recordings from each speaker to assess the degree of accented-ness using one of four labels: native, non-native (light accent), non-native (medium accent), or non-native (heavy accent). To assess inter-annotator agreement, we computed Quadratic Weighted Kappa (QWK) \cite{cohen1968weighted}. Pairwise QWK scores were 0.40 (fair agreement), 0.47 (moderate agreement), and 0.66 (substantial agreement), reflecting the subjectivity of the task. \\ %

\noindent\textbf{Human Acceptability Judgments. }
We additionally include 
utterance-level human pronunciation assessments for all recordings from two speakers in the test set (a total of 402 utterances). 
These annotations were provided by the same three native English language tutors referred to in the previous task. %
For each recording, annotators listened to the audio and judged whether the pronunciation is acceptable or not. They also provided a list of incorrectly pronounced words, a judgment on whether the pronunciation altered the meaning of the word, and optional notes explaining their decision. %
We preserve the separate judgments rather than merging them into a single label. 
The purpose of these annotations is to support analysis of cases where ASR output, intended response, and human perception may diverge. This is particularly important for accented speech, where a response may be intelligible and acceptable to human listeners despite differing from a canonical pronunciation. %

The three annotators agreed on 68\% of samples. The remaining 32\% received mixed annotations, underscoring the subjective nature of the task. Only 6\% of samples in total were judged as needing correction by all judges. In all these cases, the accented utterance was deemed to affect the meaning of the uttered word or phrase.  

\section{Dataset Structure and Splits}
CARB is released as a single dataset with three splits: train, dev, and test, each with parallel recordings by different speakers. The splits are item-disjoint; 
no context-answer pair appears across splits. They are also speaker-disjoint in evaluation sets; 
dev and test sets each have a set of unique speakers. We also have a small number of speakers that appear in all sets to test the performance for seen vs. unseen speakers (e.g., for applications where personalization is possible). 
This design supports evaluation on new items and new speakers, while still allowing analysis of both seen and unseen speakers accent conditions.

\subsection{Speaker Groups}
Speakers in CARB are organized into six recording groups according to the part of the dataset they contributed to: %

\begin{itemize}
\item \textbf{G1: Full dataset context and answer.} Two speakers recorded both contexts and answers for the whole dataset, including one native and one non-native speaker.

\item \textbf{G2: Full dataset answer.} Eight speakers recorded only the answers for the whole dataset.

\item \textbf{G3: Dev set context and answer.} One  speaker recorded both contexts and answers for the dev split.

\item \textbf{G4: Dev set answer.} 39  speakers recorded only the answers for the dev split.

\item \textbf{G5: Test set context and answer.} One  speaker recorded both contexts and answers for the test split.

\item \textbf{G6: Test set answer.} 87  speakers recorded only the answers for the test split.

\end{itemize}

The geographical coverage of the dataset is illustrated in Figure~\ref{fig:dataset_map},  summarizes the representation of different countries across splits. 

\subsection{Dataset Statistics}
The CARB dataset consists of 2,244 context-answer entries, covering a diverse speaker population spanning 55 native languages, 54 countries of origin, and 57 unique native language-country pairs. 
The dataset is gender-balanced, with 54\% female and 46\% male speakers, ranging in age from 19 to 63 years ($\mu = 33.9$, $\sigma = 9.9$). 
We summarize the size and composition of CARB at both the dataset entry and recording levels, %
as each dataset entry may be associated with multiple recordings depending on the recording type (context/answer) and speaker group. Statistics are provided in Table~\ref{tab:overall_statistics}. 

\begin{table}[t]
\centering
\caption{Overall statistics of CARB dataset.}
\label{tab:overall_statistics}
\setlength{\tabcolsep}{3pt}
\rowcolors{2}{gray!10}{white}
\begin{tabular}{lrrrr}
\toprule
\textbf{Statistic} & \textbf{Train} & \textbf{Dev} & \textbf{Test} & \textbf{Total} \\
\hline
\multicolumn{5}{c}{\textbf{Overall Dataset Statistics}}\\\hline
\#Speakers & 10 & 40 & 88 & 138 \\
\#Unique Native Languages & 8 & 24 & 50 & 57 \\
\#Unique Countries of Origin & 10 & 24 & 50 & 54 \\
\#Context-answer entries & 1,800 & 222 & 222 & 2,244 \\
\#Word Responses & 832 & 124 & 125 & 1,081 \\
\#Phrase Responses & 61 & 23 & 21 & 105 \\
\#Sentence Responses & 893 & 57 & 58 & 1,008 \\
\#Numeric Responses & 14 & 18 & 18 & 50 \\
\#Correct Responses & 1,686 & 167 & 174 & 2,027 \\
\#Incorrect Responses & 114 & 55 & 48 & 217 \\
\hline
\hiderowcolors
\multicolumn{5}{c}{\textbf{Recording-level Statistics}}\\\hline
\showrowcolors
\#Context Recordings & 3,600 & 666 & 666 & 4,932 \\
\#Answer Recordings & 19,800 & 10,878 & 21,756 & 52,434 \\
\#Total Recordings& 23,400 & 11,544 & 22,422 & 57,366 \\
Duration Context Recordings (h) & 2.2 & 0.5 & 0.6 & 3.3 \\
Duration Answer Recordings (h) & 16.3 & 8.0 & 15.5 & 39.8 \\
Duration Total Recordings (h) & 18.5 & 8.5 & 16.1 & 43.1 \\
\bottomrule
\end{tabular}
\end{table}

\subsection{Dataset Schema}
We provide structured metadata for dataset entries, speakers, and recordings, including the transcript, speaker identifier, speaker group, speaker country of origin, presence/absence of speaker's accent in train set, data split, response length and correctness, and recording type (context or response), as shown in Tables \ref{tab:carb_examples} and \ref{tab:carb_speaker_rec_example}. %
For human judgements of degrees of accented-ness and correctness, we provide individual labels for each judge rather than a single merged label. %

\section{Experiments}
In this section, we describe experimental results where we benchmark various state-of-the-art ASR models on CARB. 
The objective of these experiments is to illustrate the various use cases and analyses supported by CARB and establish initial baseline results under the conditions represented in this dataset. %
Considering the various dimensions represented in the dataset and due to space constraints, we only report a subset of possible experimental settings, focusing mainly on the individual dimensions of accent, response length, and context. We also aggregate results per geographic region and demonstrate model and uneven performance across populations.  
All results are reported on the test set.
\subsection{Models}
We conduct our evaluation using three ASR model families: 
(1) \textit{whisper-large-v3}\footnote{\href{https://huggingface.co/openai/whisper-large-v3}{openai/whisper-large-v3}}~\cite{radford2023robust}, a transformer based encoder-decoder model; (2) \textit{wav2vec2-large-960h}\footnote{\href{https://huggingface.co/facebook/wav2vec2-large-960h}{facebook/wav2vec2-large-960h}} a self-supervised wav2vec2 ASR model fine-tuned on 960 hours of LibriSpeech \cite{panayotov2015librispeech} and \textit{wav2vec2-large-960h-lv60-self}\footnote{\href{https://huggingface.co/facebook/wav2vec2-large-960h-lv60-self}{facebook/wav2vec2-large-960h-lv60-self}}~\cite{baevski2020wav2vec}, which extends wav2vec2 by using additional self-training on Libri-Light\cite{kahn2020librilight} data; and (3) NVIDIA models with fastconformer encoder and various decoder types, namely 
\textit{parakeet-tdt-0.6b-v2}\footnote{\href{https://huggingface.co/nvidia/parakeet-tdt-0.6b-v2}{nvidia/parakeet-tdt-0.6b-v2}}, \textit{canary-qwen-2.5b}\footnote{\href{https://huggingface.co/nvidia/canary-qwen-2.5b}{nvidia/canary-qwen-2.5b}}, \textit{parakeet-rnnt-1.1b}\footnote{\href{https://huggingface.co/nvidia/parakeet-rnnt-1.1b}{nvidia/parakeet-rnnt-1.1b}}, \textit{parakeet-ctc-1.1b}\footnote{\href{https://huggingface.co/nvidia/parakeet-ctc-1.1b}{nvidia/parakeet-ctc-1.1b}}, and \textit{parakeet-tdt-1.1b}\footnote{\href{https://huggingface.co/nvidia/parakeet-tdt-1.1b}{nvidia/parakeet-tdt-1.1b}}. 
The last three 1.1B Parakeet models were selected as they were trained on the same data and differ only in architecture, enabling a controlled comparison of architectural choices.

\subsection{Metrics}
We report word error rate (WER) and character error rate (CER) as the primary evaluation metrics. CER is included because many CARB items are short, and WER can be especially harsh for single-word responses. %
For contextually incorrect responses, we separately report the rate of true-to-speech transcriptions and the rate of transcriptions that are `corrected', where the model outputs the contextually correct or expected answer even when the spoken utterance is incorrect. %

\subsection{Results}
\noindent \textbf{Overall ASR Performance.} In Table~\ref{tab:overall_asr_results}, we report ASR performance on the CARB test split, for the whole set as well as answer and context items separately. Parakeet-tdt-0.6B achieves the lowest overall error rates (12.4\% WER, 5.4\% CER), narrowly ahead of Whisper (12.7\% WER, 6.16\% CER), while wav2vec2 is the weakest among the benchmarked models (33.9\% overall WER). \\

\begin{table}[tbp]
\centering
\caption{Overall ASR performance on CARB test set.}
\label{tab:overall_asr_results}
\setlength{\tabcolsep}{3pt}
\rowcolors{2}{gray!10}{white}
\begin{tabular}{lcccccc}
\toprule
\multicolumn{1}{c}{} & \multicolumn{2}{c}{\textbf{Overall}} & \multicolumn{2}{c}{\textbf{Answer}} & \multicolumn{2}{c}{\textbf{Context}} \\\cmidrule(lr){2-3}\cmidrule(lr){4-5}\cmidrule(lr){6-7}
\hiderowcolors
\multicolumn{1}{c}{\textbf{Model}} & \multicolumn{1}{c}{\textbf{WER}} & \multicolumn{1}{c}{\textbf{CER}} & \multicolumn{1}{c}{\textbf{WER}} & \multicolumn{1}{c}{\textbf{CER}} & \multicolumn{1}{c}{\textbf{WER}} & \multicolumn{1}{c}{\textbf{CER}} \\\hline
\showrowcolors
whisper-large-v3     & 0.13 & 0.06 & \textbf{0.13} & 0.07 & \textbf{0.03} & \textbf{0.02} \\
parakeet\_tdt\_0.6b  & \textbf{0.12} & \textbf{0.05} & \textbf{0.13} & \textbf{0.06} & 0.04 & \textbf{0.02} \\
parakeet\_rnnt\_1.1b & 0.15 & 0.08 & 0.16 & 0.09 & 0.04 & \textbf{0.02} \\
parakeet\_ctc\_1.1b  & 0.20 & 0.11 & 0.21 & 0.12 & 0.05 & 0.03 \\
parakeet\_tdt\_1.1b  & 0.14 & 0.07 & 0.15 & 0.07 & 0.04 & \textbf{0.02} \\
canary-qwen-2.5b     & 0.16 & 0.07 & 0.17 & 0.08 & 0.05 & 0.03 \\
wav2vec2-large-self  & 0.24 & 0.11 & 0.25 & 0.11 & 0.12 & 0.05 \\
wav2vec2-large       & 0.34 & 0.16 & 0.35 & 0.16 & 0.21 & 0.09 \\

\bottomrule
\end{tabular}
\end{table}

\begin{figure}[tbp]
    \centering
    \includegraphics[width=0.8\linewidth]{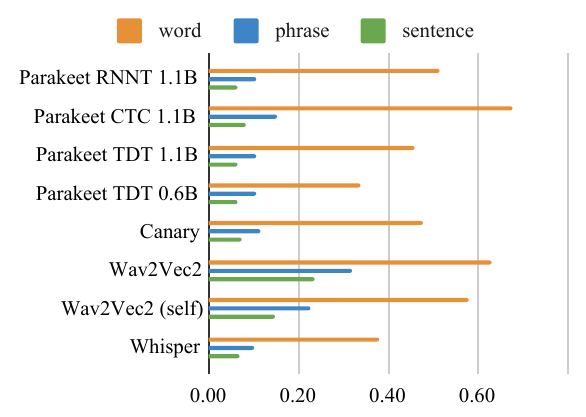}
    \caption{Performance of models across response types.}
    \label{fig:results_respone_type}
\end{figure}

\noindent\textbf{Effect of Response Length}
Overall, Off-the-shelf models show higher error rates for answers compared to context utterances as seen in Table \ref{tab:overall_asr_results}, reflecting the challenging nature of transcribing responses without context, as described in the introduction.  %
Figure~\ref{fig:results_respone_type} further breaks down the performance on Answer utterances by response length. Single-word responses produce the highest error rates, with WER ranging between 34-68\%, which is likely because they provide limited internal linguistic context. WER drops significantly for phrases (10-32\%) and sentences (7-24\%). \\%

\begin{figure}[tbp]
    \centering
    \includegraphics[width=\linewidth]{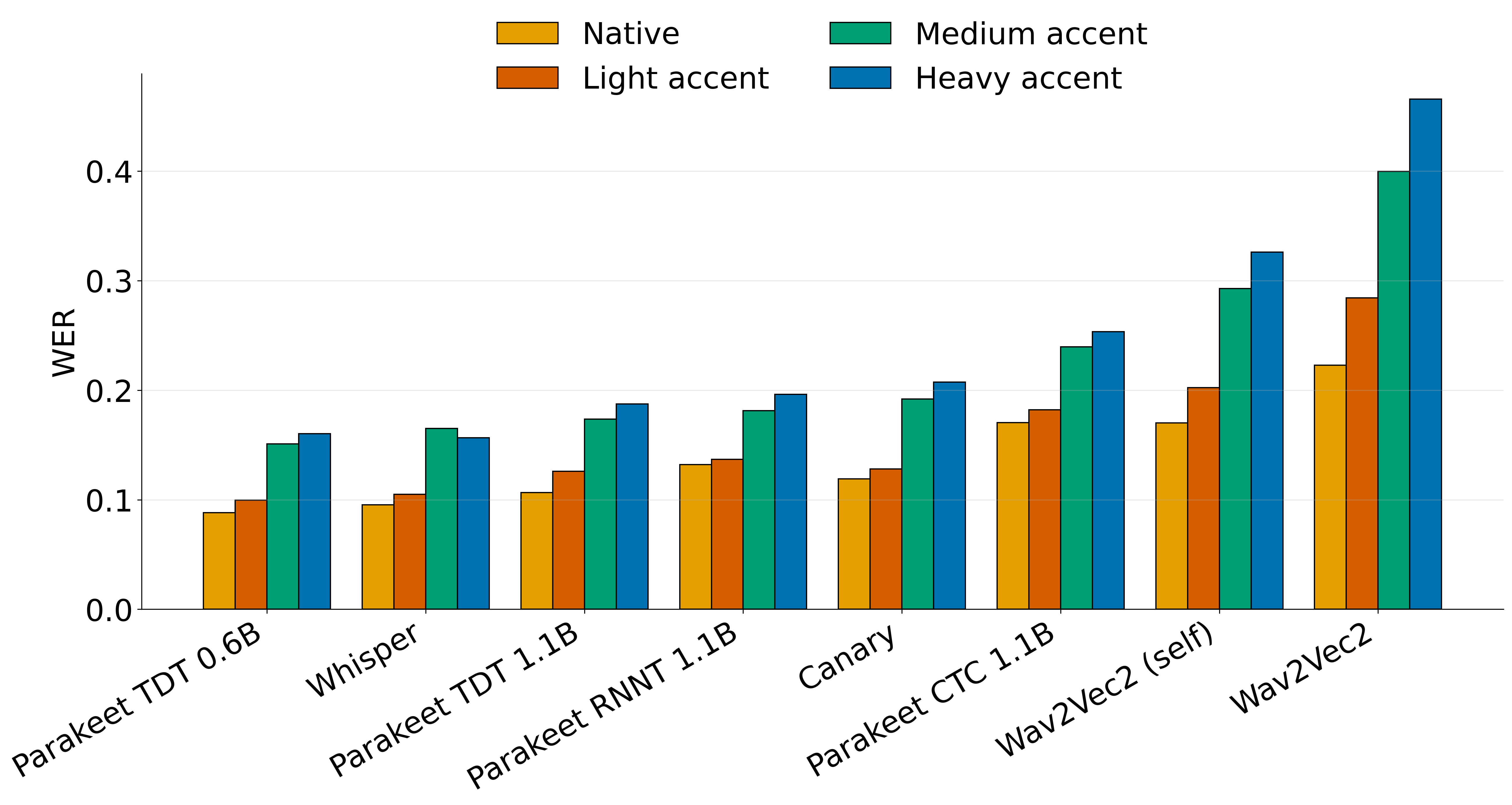}
    \caption{Performance of models across 
    different degrees of perceived accent.}
    \label{fig:results_accents}
\end{figure}

\noindent\textbf{Performance Across Accents. }
\begin{figure}[tbp]
    \centering
    \includegraphics[width=\linewidth]{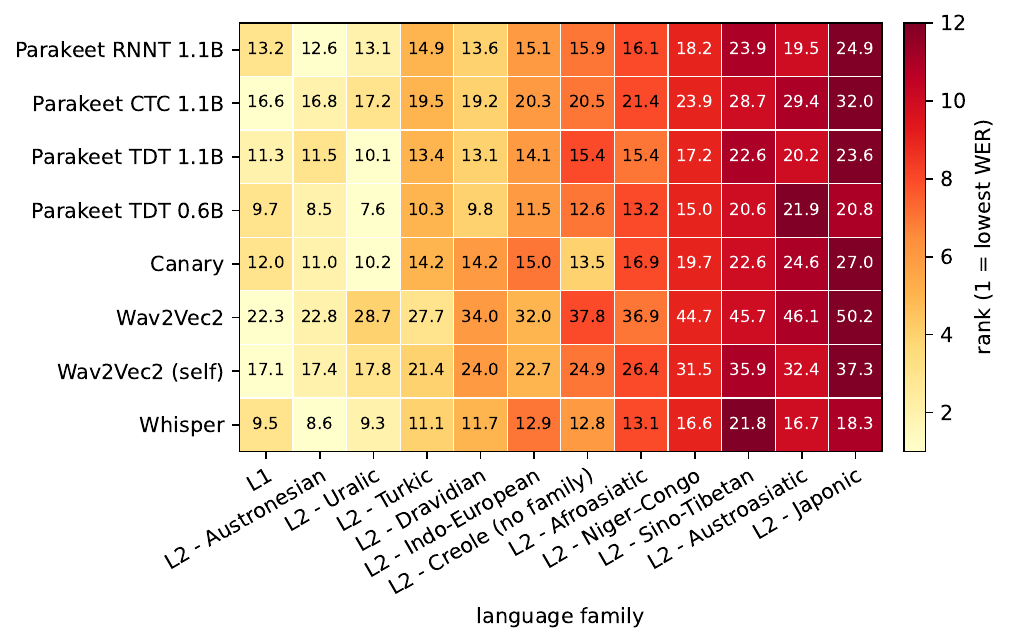}
    \caption{%
    Performance of models (WER) across L1 English speakers and L2 speakers grouped by the language family of their first language. Cell colors indicate the ranking of speaker groups for each model, where the lowest WER is ranked 1.
    }
    \label{fig:results_accents_ranking}
\end{figure}
In Figure~\ref{fig:results_accents_ranking}, we show the performance of ASR models across L1 English speakers and L2 speakers grouped according to the language family of their first language. 
Japonic, Sino-Tibetan, and Niger--Congo are the most challenging groups across all models, whereas performance on English-L1 and Austronesian speakers is consistently better.
We also investigate ASR performance across varying degrees of speakers' accented-ness as judged by the human annotators. As shown in Figure~\ref{fig:results_accents}, WER increases proportional to accent strength. On average, WER roughly doubles (ratios ranging from ~1.4 to ~2.3) when moving from native to heavy-accented speech. 
Analyzing results against human acceptability scores, this bias is not justified by pronunciation errors, but stems purely from the effect of accent. Out of all accented responses that are judged as acceptable by the human tutors, about 40\% of them include ASR errors using Whisper (best), 48-68\% using NVIDIA models, and 71-80\% using wav2vec2 models (worst). This demonstrates how ASR systems incorporated into educational applications could potentially result in systematic biases when deployed for diverse or under-represented groups.   %
\\

\begin{figure}[tpb]
    \centering
    \includegraphics[width=\linewidth]{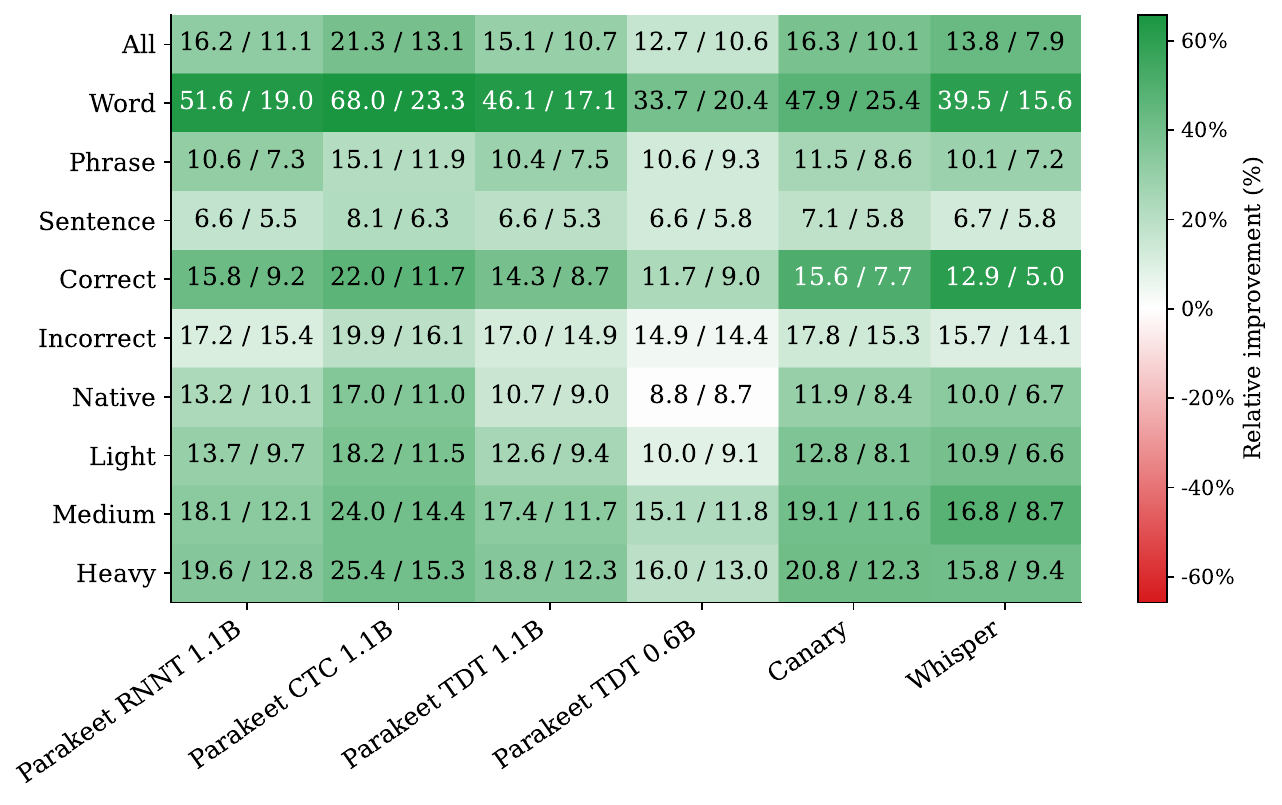}
    \caption{%
    Heatmap showing WER with (W) and without (WO) context. Each cell shows WO / W WER, across different categories. The cell color indicates the relative improvement achieved by using context (green indicates better performance with context).
    }
    \label{fig:results_context_impact}
\end{figure}

\begin{figure}[tpb]
    \centering
    
\includegraphics[width=\linewidth]{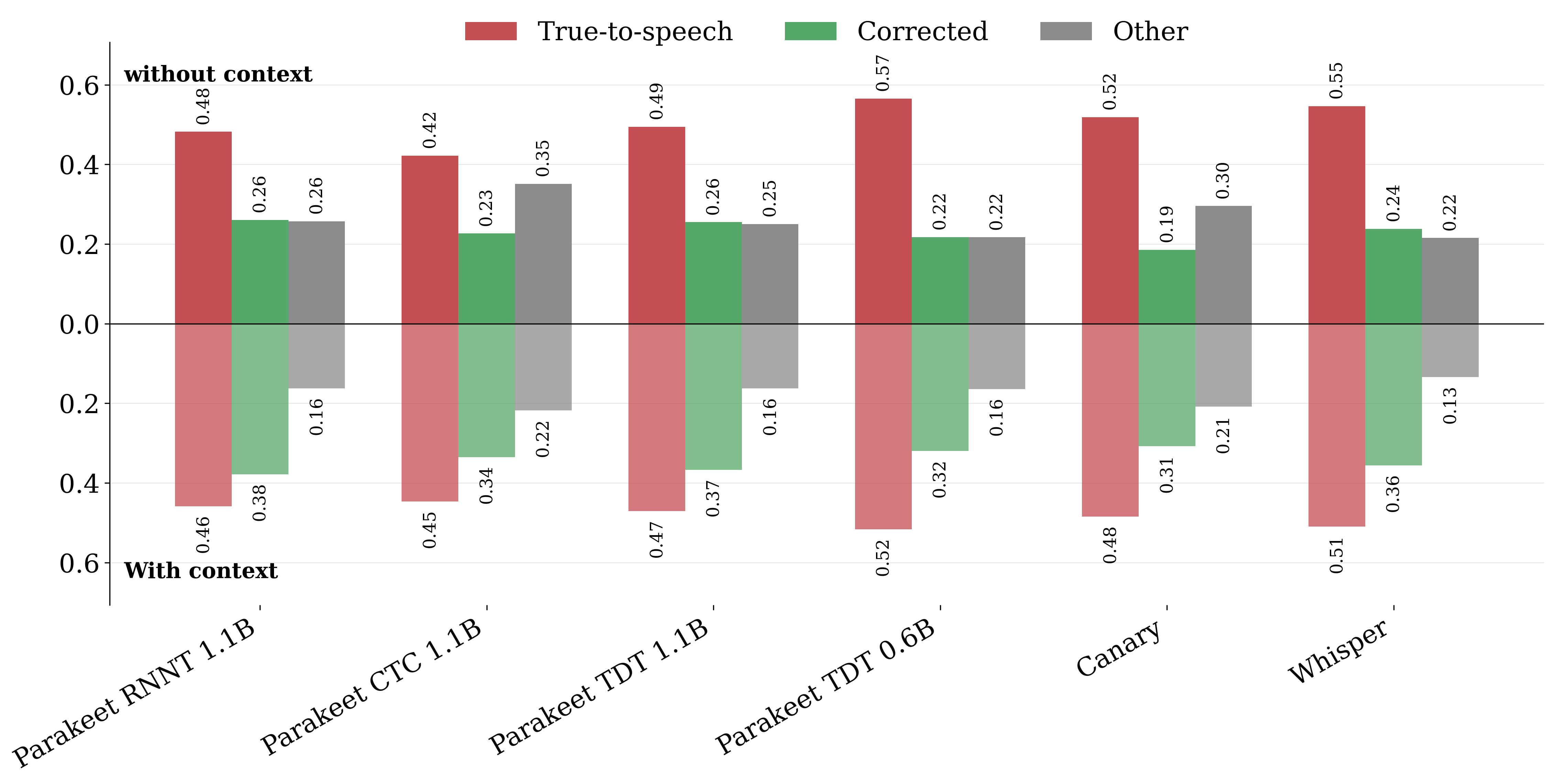}
\caption{Analysis of transcription of incorrect responses, with and without context. We show the rate of transcriptions that are true to the spoken content, and the rate of corrected transcripts. }\label{fig:analysis_incorrect}
\end{figure}

\noindent\textbf{Effect of Context. }
To evaluate the effect of context, we concatenate each answer recording with the corresponding context audio to form a single audio segment. This segment is then transcribed, after which we do alignment search to retain only the answer words. The resulting sliced transcript is treated as the with-context hypothesis. For comparison, we also transcribe the answer audio alone, producing the without-context hypothesis. %
The full results for a subset of models are plotted in Figure~\ref{fig:results_context_impact}. Adding context generally cuts overall WER by 2 - 8 \% for all models, with the biggest gains on single-word answers and medium/heavy accents. The relative improvements are smallest when the answers are incorrect, for which incorporating the context does not provide useful semantic biasing. 
We further analyze system behavior on responses that are incorrect given the context. We calculate how often the model outputs the words as spoken (true-to-speech) vs. how often it corrects the answer inspite of the acoustic evidence. The results are shown in Figure \ref{fig:analysis_incorrect}. We note here that, while overall error rates are improved with context, all models tends to over-correct the responses when context is provided, compared to the case without context, which reflects the complex influence of context on recognition performance. 

\section{Conclusion}
This paper presented a new dataset designed to benchmark ASR model for complex use-cases, where accent variations, context, response length, and phonetic ambiguity, may have unexpected impacts on performance. For instance, 
in real-world ASR deployment, the degree to which the model should utilize the context for prediction depends on the use case. 
In settings such as keyword spotting or voice commands, it may be desirable for the system to bias its prediction toward the expected response. In language learning and pronunciation assessment, on the other hand, context could potentially mask user errors and compromise the intended learning process. CARB was designed to analyze this tension while also evaluating the impact of natural accent variations and utterance length. %
Considering the subjective nature of the task, we also provide human judgments that help distinguish between ASR errors and acceptable accented pronunciations. 
We benchmarked various state-of-the-art models, and demonstrated how the dataset can be used to analyze these factors in a controlled manner. The findings presented in the paper underscore the failure modes of state-of-the-art ASR models, including demographic bias, performance degradation in the absence of context, and the tendency to over-correct in the presence of context. The dataset includes a training split that can be used in future work to study the impact of fine-tuning on generalization, be it to different speakers, accent groups, response types, or context. %

\bibliographystyle{IEEEtran}
\bibliography{mybib}

\end{document}